\documentclass[10pt,letterpaper]{article}

\usepackage{cogsci}

\cogscifinalcopy 

\usepackage{pslatex}
\usepackage{apacite}
\usepackage{float} 

\usepackage{colortbl}
\usepackage{tikz}
\usepackage{booktabs}
\usepackage{tabularx}
\usepackage{graphicx}
\usepackage{multirow}
\graphicspath{{acl-ijcnlp2021-templates/images}}
\usepackage{subcaption}
\usepackage{xurl}

\usepackage[normalem]{ulem}

\newcommand{\winogender}[0]{Winogender}
\newcommand{\winobias}[0]{WinoBias}
\newcommand{\sys}[1]{\textit{System #1}}
\newcommand{\kirst}[0]{\emph{s2e}}
\newcommand{\as}[0]{anti-stereotypical}
\newcommand{\ps}[0]{pro-stereotypical}
\newcommand{\comment}[1]{}

\title{Comparing Humans and Models on a Similar Scale:\\
Towards Cognitive Gender Bias Evaluation in Coreference Resolution}
 
\author{{\large \bf Gili Lior (gili.lior@mail.huji.ac.il)} \\
  School of Computer Science and Engineering, \\ The Hebrew University of Jerusalem, Israel
  \AND {\large \bf Gabriel Stanovsky (gabriel.stanovsky@mail.huji.ac.il)} \\
  School of Computer Science and Engineering, \\ The Hebrew University of Jerusalem, Israel}

\begin{document}

\maketitle

\begin{abstract}

Spurious correlations were found to be an important factor explaining model performance in various NLP tasks (e.g., gender or racial artifacts), often considered to be ``shortcuts'' to the actual task.
However, humans tend to similarly make quick (and sometimes wrong) predictions based on  societal and cognitive presuppositions.
In this work we address the question: 
\textit{can we quantify the extent to which model biases reflect human behaviour?}
Answering this question will help shed light on model performance and provide meaningful comparisons against humans. 
We approach this question through the lens of the \textit{dual-process theory} for human decision-making. This theory differentiates between an automatic unconscious (and sometimes biased) ``fast system'' and a ``slow system'', which when triggered may revisit earlier automatic reactions. 
We make several observations from two crowdsourcing experiments of gender bias in coreference resolution, using
self-paced reading 
to study the ``fast'' system, and question answering to study the ``slow'' system under a constrained time setting.
On real-world data humans make $\sim$3\% more gender-biased decisions compared to models, while on synthetic data models are $\sim$12\% more biased. We make all our of our code and data publicly available.\footnote{\url{https://github.com/SLAB-NLP/Cog-GB-Eval}}

\end{abstract}

\noindent \textit{\textbf{Disclaimer:} This is an extended version of a paper appearing in CogSci 2023. See Appendix for a detailed changelog between the two papers.}

\section{Introduction}
Recent work has identified that large-scale models achieve impressive results on various natural language benchmarks by exploiting correlations which do not seem semantically meaningful for solving the task~\cite{gururangan-etal-2018-annotation,gardner-etal-2021-competency}.
Leveraging such \emph{spurious correlations} is often considered an indication that models do not solve the actual task, but instead resort to finding statistical ``shortcuts'' around the problem~\cite{geva-etal-2021-aristotle,savoldi-etal-2021-gender}.


\begin{figure}[t!]
    \centering
    \includegraphics[width=0.6\columnwidth]{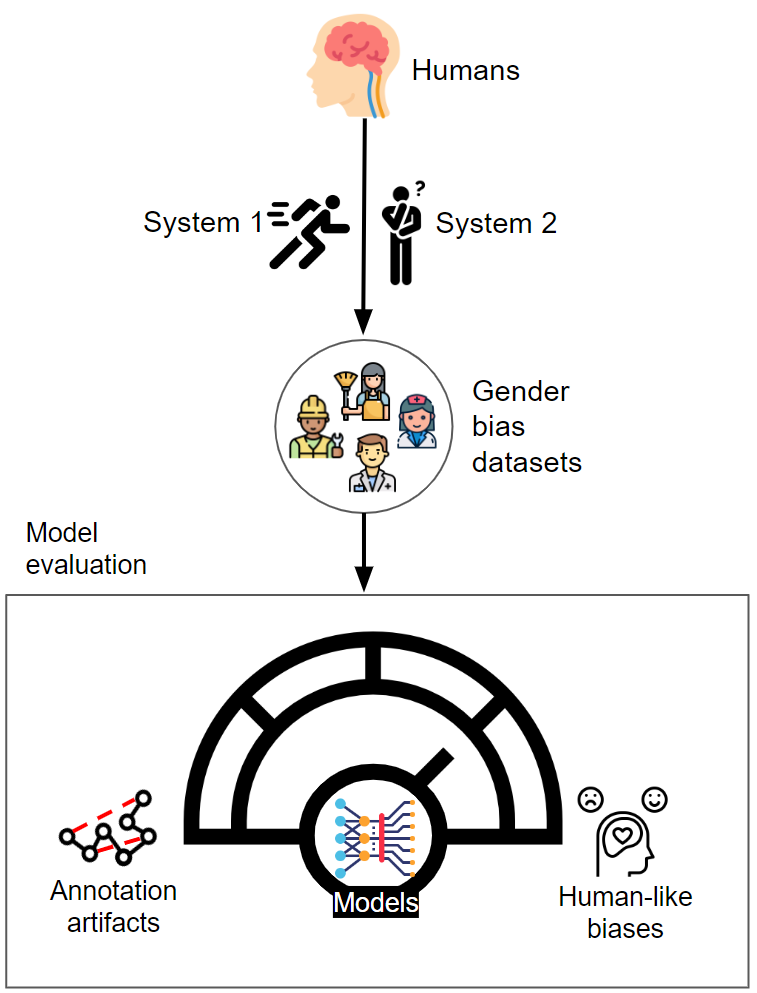}
    \caption{High-level overview of our work. 
    We develop an evaluation paradigm for human subjects following the dual-process theory for decision making and compare them to model biases, analyzing the difference between real-world and synthetic sentences.
    }
    \label{fig:fig1}
\end{figure}

In parallel, works in cognitive psychology identify that finding shortcuts may in fact be a  \textit{feature} of human intelligence, 
which on the one hand helps us cope with missing or implicit information~\cite{grice1975logic,grice1989studies}, while on the other hand may also lead to harmful behavior.
In the context of gender bias in coreference resolution, which will be the focus of our work, studies have found that human subjects tend to prefer the stereotypical reading in various modalities, such as event-related brain potentials, reading times, or eye movements~\cite{osterhout1997brain,kennison2003comprehending,duffy2004violating}.

In this work we propose to integrate findings from these two lines of research and quantify the extent to which model biases resemble human behavior.
 We distinguish between two ends of a spectrum, 
 as shown in Figure~\ref{fig:fig1}. 
 On the one hand we place \emph{annotation artifacts}, which hold only in specific training sets, e.g., associating the word ``cat'' with contradiction in NLI~\cite{gururangan-etal-2018-annotation}.
 On the other hand of the spectrum we place \emph{human-like biases} which are sometimes useful in real-world scenarios (e.g., in common sense reasoning~\cite{lent-sogaard-2021-common}), but also produce harmful, unwanted behavior (as in gender bias~\cite{schwartz-stanovsky-2022-limitations}). These are likely to arise in any real-world dataset, and may require subtle debiasing techniques in either modelling or data collection.

To place model biases on this spectrum, we develop human annotation interfaces and derive evaluation metrics which compare between humans and models, thus putting them on the same scale.
In particular, we focus on gender bias in coreference resolution in the English language,  which was widely studied in machine learning and psycholinguistics, allowing us to explore results in the intersection of these areas.

To achieve this, we study human biases through the lens of the \textit{dual-process theory}~\cite{evans2008dual},
which posits that there are two cognitive systems participating in humans' decision making process. \sys{1} is fast, associative and automatic, while \sys{2} is slow, conscious and effortful. 
\sys{1} heuristics are considered a survival mechanism. Humans make thousands of decisions a day, and if all of them were consciously processed, our brain would not handle the cognitive load. But on the other hand, when \sys{1} ``shortcuts'' are wrong and \sys{2} does not revise it, erroneous and biased decisions may occur~\cite{kahneman2011thinking}. 

Within this framework, we propose two human experiments to quantify the heuristics made by \sys{1}. 
The first experiment tests \sys{1} directly, by examining how gender bias manifests in self-paced reading~\cite{jegerski2013self}, which approximate eye tracking, largely considered to be an unconscious process~\cite{rayner1998eye}.
The second experiment is question answering (QA) over coreference-related questions. QA is likely to invoke \sys{2}, as it requires more conscious effort~\cite{wang2003cognitive}. We then add different artificial time constraints,
to examine how \sys{1} heuristics are expressed in a task that requires more cognitive effort.


Finally, we crowdsource annotations for the two experiments over synthetic and real-world sentences, and make several important observations, comparing humans to two state-of-the-art coreference models.
Both experiments surface comparable gender biases to those shown by models. Specifically, in the QA experiment over the natural sentences, models' overall accuracy is significantly lower than humans, but both show similar biases. In contrast, for the synthetic sentences, the models' overall accuracy was closer to humans, but models have shown  larger gender bias.

To the best of our knowledge our work presents a first quantitative evaluation of gender bias in coreference resolution models versus human behavior, specifying the conditions needed to elicit comparable biases from humans through time constraints.
Our results indicate that model biases indeed resemble decisions made by humans with restricted attention span. Future work may leverage our evaluation paradigm and revisit it for other tasks and future models. 

\section{Background}

We begin by describing previously published datasets designed to test model biases in coreference resolution 
. To measure human performance, 
we then discuss Maze~\cite{forster2009maze}, a self-paced reading approach approximating eye-tracking measurements.

\subsection{Gender Bias Datasets}\label{sec:datasets-backgroud}

We use three coreference gender bias datasets as outlined below, and summarized in Table~\ref{tab:datasets}.
\begin{figure*}
\centering
\begin{subfigure}{.4\textwidth}
  \includegraphics[width=\textwidth]{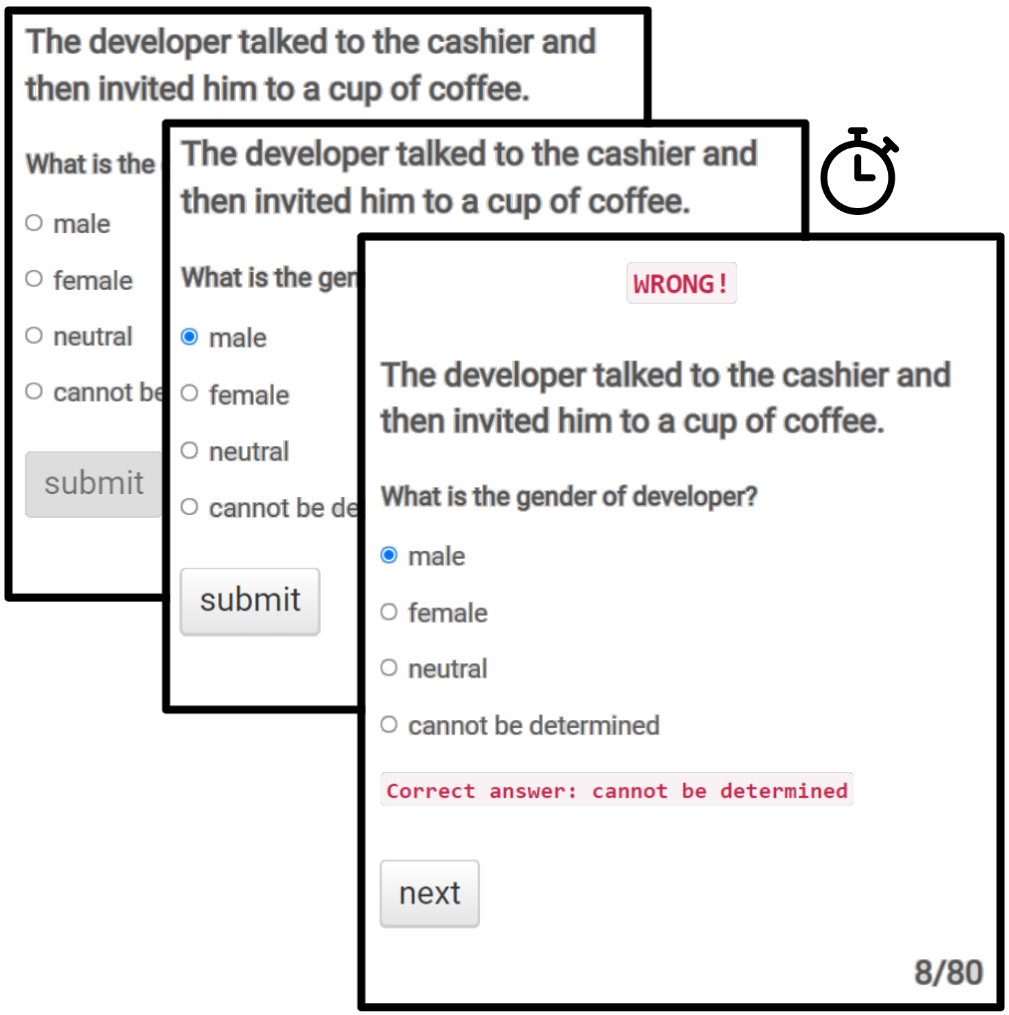}
  \caption{QA calibration interface. A sentence and a question are shown for an unlimited time. 
  After submitting the answer, a feedback is shown on screen, including the correct answer. We record the participants choice and the time they took to answer.}
  \label{fig:QA-cal}
\end{subfigure}\hspace{15mm}
\begin{subfigure}{.4\textwidth}
  \includegraphics[width=\textwidth]{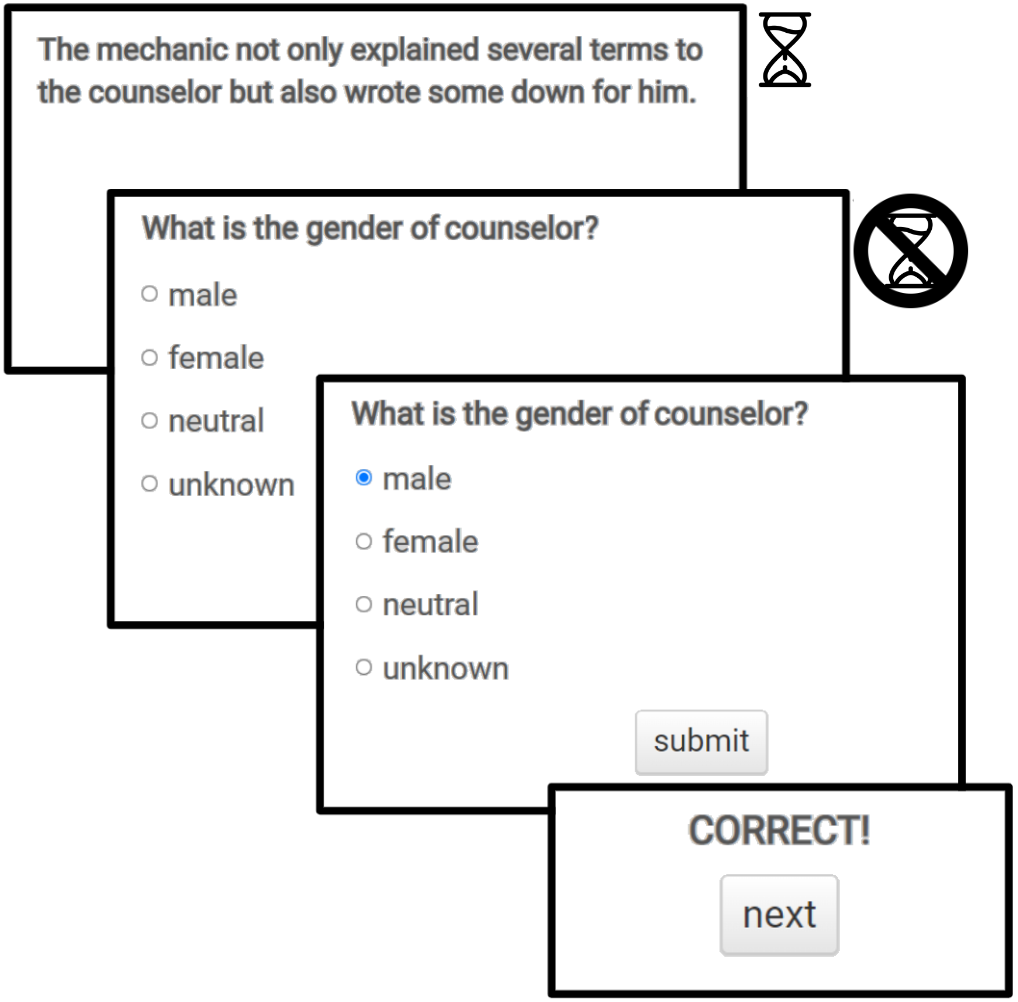}
  \caption{QA Experiment interface. A sentence is shown for a limited time. Then, the sentence disappears and only a question is shown, for an unlimited time. After submitting an answer, a feedback message is shown. Here we only record the the participants choice.}
  \label{fig:QA-exp}
\end{subfigure}
\caption{QA calibration and main experiment interfaces.}
\label{fig:QA-flow}
\end{figure*}

\winobias{}~\cite{zhao-etal-2018-gender} and \winogender{}~\cite{rudinger-etal-2018-gender} consist of 3,888 synthetic, short sentences. Each of the sentences conforms to a similar template consisting of two entities, identified by their profession, and a single referring pronoun. The datasets are balanced with respect to stereotypical gender-role assignment (e.g., female secretaries) versus non-stereotypical assignment (e.g., male nurses). 
These datasets are good for controlled experiments but consist of a small variety of linguistic constructions, and do not represent real-world distributions. 

In contrast, the BUG corpus~\cite{levy-etal-2021-collecting-large} aims to find such templates ``in the wild''. It consists of 1,720 sentences sampled from natural corpora (e.g., Wikipedia and PubMed) and better approximates real-world distribution in terms of sentence length, vocabulary and gender-role stereotypes. 
Similar to \winogender{} and \winobias{}, each sentence in BUG presents entities identified via their profession and a referring pronoun. 
BUG also provides a binary annotation for each sentence marking whether is conforms to societal norms.
For accuracy sake, we use a subset of BUG which was manually annotated.
\begin{table}
\centering
\resizebox{0.92\columnwidth}{!}{%
\begin{tabular}{l|rr|rrrr }
\toprule
                           & \multicolumn{2}{c|}{Original} & \multicolumn{2}{c}{QA} & \multicolumn{2}{c}{MAZE} \\
                           & \#pro           & \#anti          & \#pro        & \#anti       & \#pro        & \#anti        \\ \midrule
\winobias   & 1582          & 1586          & 756        & 717        & 607        & 603         \\
\winogender & 216           & 216           & 203        & 216        & 35         & 35          \\
BUG                        & 865           & 420           & 431        & 271        & 565        & 315         \\ \bottomrule
\end{tabular}%
}
\caption{Statistics for coreference gender bias datasets. ``Original'' presents the number of sentences in each of the datasets. ``QA'' and ``MAZE'' show the number of sentences in our experiments, further decomposed into  \ps{} and \as{} sentences.
The reduction in sampling sizes is due to additional filtering and distribution tuning. See the Experiments section for more details.}
\label{tab:datasets}
\end{table}


\subsection{Maze}
For our proposed evaluation metric presented in the Experiments section,
we use Maze~\cite{forster2009maze}, a platform for measuring self-paced reading~\cite{jegerski2013self}. This platform is an alternative for eye-tracking measurements~\cite{witzel2012comparisons}, that does not require specialized equipment and in-house annotators. Instead, Maze can be easily deployed on crowdsourcing platforms, allowing us to collect annotations at scale.

As exemplified in Figure~\ref{fig:MAZE-flow}, Maze iteratively presents two options for the next word in a sentence, and a human annotator needs to select the  most probable alternative given previously seen words. The time for choosing the correct word approximates its reading time.




\section{Working Definitions}
\label{sec:key-definitions}

In this section, we formally define key concepts commonly used throughout the paper.

\textbf{Gender.} We use existing gender bias corpora, as described in the Background section,
using pronouns with three grammatical genders: feminine, masculine, and neutral. 
The complete list of pronouns and their distribution in these corpora is shown in Table~\ref{tab:grammatical-gender-pronouns} in the Appendix. 
These datasets are generally devoid of other types of pronouns, such as neo-pronouns.\footnote{\url{https://www.unf.edu/lgbtqcenter/Pronouns.aspx}}  Collecting corpora for diverse types of pronouns is left as an important avenue for future work, e.g., as outlined by \cite{lauscher2022welcome}.

\textbf{Pro-stereotype/Anti-stereotype.}\label{par:pro-anti-def} 
A coreference relation between a pronoun and an entity in a sentence is deemed pro-stereotypical if the referring pronoun's  gender conforms to societal norms (e.g., ``nurse'' and ``she''), otherwise it is marked anti-stereotypical (e.g., ``cleaner'' and ``he'').
These definitions naturally extend to sentences with a single pronoun. These are deemed pro or anti stereotypical according to the relation between the entity and its referring pronoun. 
To estimate the stereotypical gender norm per profession we use labels provided in the previously-published gender bias datasets~\cite{rudinger-etal-2018-gender,zhao-etal-2018-gender}, based on both human annotations and reports published by The U.S. Bureau of Labor Statistics.\footnote{We note that these norms may vary between cultures. Adapting them to other communities is left for an important future work.}


\textbf{Gender bias.}\label{par:gender-bias-def} 
We adopt the \emph{Historical Bias} definition~\cite{mehrabi2021survey}: \emph{``the already existing bias and socio-technical issues in the world and can seep into from the data generation process even given a perfect sampling and feature selection''}.
This definition connects between the physical world and how it manifests in the the training data. 
In particular, historical bias appears when models make predictions based on the gender distribution in the training data, rather than the relations between entities in the sentence. 



\section{Experiments}
\label{sec:experiments}

In the following section  we present our two experiments for measuring human biases. The design choices we make follow common practices in psycholinguistic literature.

We choose the data for the experiments by following the \winobias{} and \winogender{} categorization into \ps{} and \as{} instances.
In addition, we add 216 sentences (5.6\%)  of those originally marked as neutral. 
Particularly, sentences in which the coreference link contradicts societal norms are considered anti-stereotypical, but also sentences in which the coreference link is to a neutral entity, yet the distractor entity is correlated with the pronoun. Consider for example, the sentence: ``The teenager confided in the therapist because she was seeking emotional support.'' 
Although ``she'' refers to the gender-neutral ``teenager'', the stereotypically feminine distractor (``therapist'') poses a pitfall for biased decisions, and is thus considered anti-stereotypical. See the full details in Table~\ref{tab:wino-types} in the Appendix.

Finally, for the BUG dataset, we ensure that our sample is balanced across professions. Table~\ref{tab:datasets} shows this  especially affected BUG, as it over-represents certain entities (e.g., ``patient'' or ``doctor'' in the PubMed corpus).

\subsection{QA Experiment}\label{sec:QA-experiment}

In this experiment we present a sentence followed by a multiple-choice question regarding the gender of an entity in the sentence, eliciting coreference resolution decisions. For example, given the sentence ``The developer talked to the cashier and invited him for a cup of coffee'', and the question: ``What is the gender of the cashier?'', the four possible answers are `male', `female', `neutral' and `unknown', and the expected answer is `male'.

QA is likely to invoke \sys{2} as it involves conscious decision making. To test \sys{2} under a constrained setting, annotators observe the sentence for a limited time before it disappears and then they can answer the question.
 See Figure~\ref{fig:QA-flow} for an example annotation interface.

\textbf{Filler questions.}
Following common practice in human annotation tasks, we introduce filler questions to prevent participants from focusing on certain aspects of the sentence (e.g., its pronoun).
We automatically formulate questions on predicate-argument relations using a pretrained QA-SRL model~\cite{fitzgerald-etal-2018-large}, that produces different question formats, e.g., asking about the subject (``who might be talking?''), object (``who was being hired?'') and other entities in the sentence.
Similarly to other psycholinguistics works, our filler questions constitute $50\%$ of the total questions in the experiment~\cite{witzel2012comparisons, kim2019testing, boyce2020maze}.

\textbf{Calibration.} 
To account for different reading paces~\cite{doi:10.1177/1529100615623267}, we begin with calibrating a baseline reading pace for each participant. 
We present the sentence for an unlimited time along with the question and measure the time it takes the participant to submit the correct answer. This is then normalized by the sentence length (in words) to approximate a participant's reading pace.
See Figure~\ref{fig:QA-cal} for an example of this interface. 

\textbf{Annotation interface.}
Each participant observes a single sentence for a limited amount of time. Then, the sentence disappears and a question regarding one of the entities in the sentence is shown for an unlimited time. 
The time each sentence is presented on screen is calculated by 
$(\alpha\cdot avg\cdot l)$,
where $avg$ is the participant's reading pace, $l$ is the length of the sentence in words, and  $\alpha$ is sampled i.i.d from $\{0.25,0.5,0.75\}$ to present the sentence for a fraction of the participant's pace. An example of this interface is shown in Figure~\ref{fig:QA-exp}. 

\textbf{Annotator feedback.} Following \cite{malmaud-etal-2020-bridging}, we show participants a feedback message indicating if they were correct after every submitted answer, both for filler questions and for the actual task. Feedback in multiple-choice questions has been shown to improve performance and reduce low-quality annotations~\cite{butler2008feedback}. To mitigate the risk of affecting responses in unintended ways, we use filler questions that prevent annotators from overspecializing in the task.

\textbf{Filtering non-coreference errors.}
This setup may produce errors which do not relate to coreference. 
For example, answering that the gender of the entity is masculine while the presented pronoun is feminine (and vice versa) does not indicate a coreference error, and is therefore ignored. 

\subsection{Self-Paced Reading Experiment}\label{sec:maze-experiment}
\begin{figure}
\centering
\includegraphics[width=0.6\columnwidth]{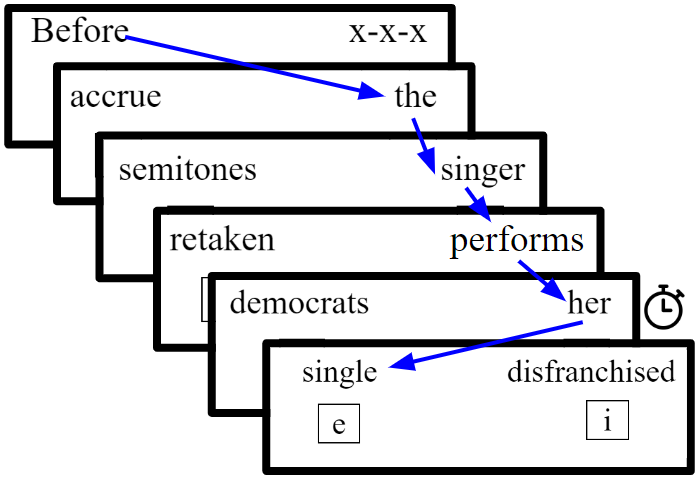}
\caption{MAZE experiment interface. At each step, participants need to distinguish the next word from a distractor, by pressing the correct keyboard key. We record the time they took for identifying the pronoun.}
\label{fig:MAZE-flow}
\end{figure}

In the second experiment, we approximate trends in reading time of pronouns in pro-stereotypical versus anti-stereotypical instances, which is  considered an unconscious process,  and hence a good proxy for \sys{1}'s  biases~\cite{rayner2009eye}.


We use MAZE to approximate the time it takes a participant to choose the pronoun in our sentences (see Figure~\ref{fig:MAZE-flow}). This implicitly measures the timing of a coreference decision since the pronoun indicates the gender of a previously mentioned entity. 
Previous work has identified that self-paced reading is a good proxy for natural reading when comparing between readings of different sentences, albeit it may overestimate the absolute reading times~\cite{yan2020expectation}.
This makes self-paced reading adequate for our purposes, as we are interested in the \textit{trends} shown in response time between \ps{} and \as{} instances. 



\textbf{Filtering ambiguous instances.}
Since MAZE presents the words in a linear order, we note that there are instances when the pronoun appears before the context needed to infer its antecedent. E.g., when reading the prefix ``The sheriff questioned the housekeeper as \emph{she}...'' it is yet unclear whether ``she'' refers to the sheriff (e.g., as in ``...\emph{she} needed to find the thief.'') or the housekeeper (e.g., in ``... \emph{she} was cleaning'').
Since the reader cannot know which of the suffixes will follow, these instances do not reflect gender bias decisions.
To address this issue in \winobias{} and \winogender{}, we sample only sentences where the pronoun appears after all verbs in the sentence, e.g., 
``The tailor thought the janitor could be good at sewing and encouraged her''. 
In a preliminary analysis we find that this heuristic may be over-strict, but leads to high precision, which was most important for our analyses.
 From BUG we sampled only sentences where the pronoun appeared after its antecedent. We find that this sampling works well for the sentences in BUG, which usually consist  of a single entity.
 
For the synthetic sentences, this sampling produces a subset of 1,335 viable sentences. Most of the instances which were filtered out come from \winogender, because in most of its sentences the pronoun appears before one of the verbs in the sentence. For BUG, this sampling produces a subset of 1,603 viable sentences.

\textbf{Annotation interface.}
At each time step participants are shown two possible words, and they need to choose the next word in the sentence according to previous context. See Figure~\ref{fig:MAZE-flow} an example. 
We allow participants to retry in case of an error, and record the time until their first answer, as well as the total time until the correct option was chosen. 


\textbf{Simulating arbitrary time limitations.}
Similarly to the QA experiment, we would like to introduce a notion of time constraints.
If we would have limited the amount of time given to distinguish the next word, a participant would either: (a) choose correctly (b) choose incorrectly (c) not respond in time. Instead of testing participants over different discrete time limitations, we make the following assumption: if a participant's response time for a correct annotation was $x$ ms, any time limit below $x$ ms would not be enough time for responding (option (c) above). 
Following, we do not limit participants reading time, but instead compute a cumulative distribution function over all possible observed response times.
Finally, we use A-Maze~\cite{boyce2020maze} to automatically generate probable distractors.

\begin{table}
\centering
\resizebox{0.9\columnwidth}{!}{%
\begin{tabular}{@{}ll|rrr|rrr@{}}
\toprule
\multicolumn{2}{l|}{\multirow{2}{*}{}}                                            & \multicolumn{3}{c|}{Wino} & \multicolumn{3}{c}{BUG} \\
\multicolumn{2}{l|}{}              & pro  & anti & \small{\textbf{$\Delta_{QA}$}} & pro  & anti & \small{\textbf{$\Delta_{QA}$}} \\ \midrule
\multirow{3}{*}{\begin{tabular}[c]{@{}l@{}}Humans\\ baseline\end{tabular}} & 0.75 & 89.8          & 89.7         & \textbf{0.1}         & 87.8   & 82.8   & \textbf{5.0}   \\
                        & 0.50     & 88.8 & 87.6 & \textbf{1.2}   & 85.7 & 79.1 & \textbf{6.5}   \\
                        & 0.25     & 78.7 & 78.5 & \textbf{0.2}   & 82.6 & 75.7 & \textbf{6.9}   \\ \midrule
\multirow{2}{*}{Models} & \small{SpanBERT} & 86.5 & 73.4 & \textbf{13.1}  & 62.2 & 60.0 & \textbf{2.2}   \\
                        & \kirst{}    & 91.3 & 77.7 & \textbf{13.6}  & 61.7 & 59.3 & \textbf{2.4}   \\ \bottomrule
\end{tabular}%
}
\caption{Human and model results in the QA task on the same sentences. `pro' and `anti' columns show results on \ps{} and \as{} gender questions. $\Delta_{QA}$ stands for the difference between the two categories (pro minus anti), indicating biased performance, approximating \sys{2} biases.}
\label{tab:QA-results}
\end{table}

\begin{figure}
\centering
\includegraphics[width=0.92\columnwidth]{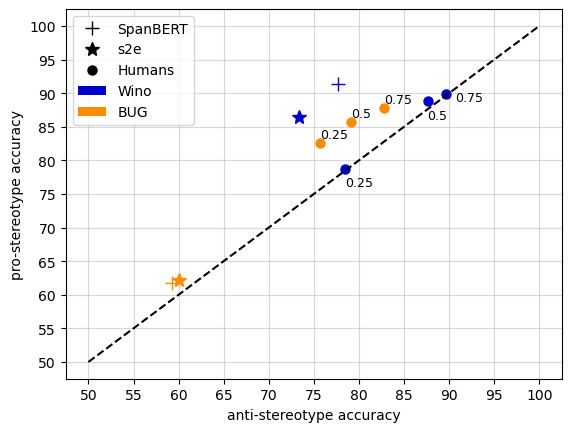}
\caption{Visualization for the results shown in Table~\ref{tab:QA-results}. The x-axis is the performance over the \as{} sentences, and the y-axis is the performance over the \ps{} sentences. Values above the dashed black line show gender biased performance. Datasets are represented by color, while humans are distinguished from models by the indicator's of your shape. $\{0.25,0.5,0.75\}$ are the fractions of the baseline reading pace given to humans. All evaluations found some degree of gender bias.}
\label{fig:QA-results}
\end{figure}

\section{Results}
\label{sec:results}



\begin{figure*}
\centering
\begin{subfigure}{0.32\textwidth}
  \includegraphics[width=\textwidth]{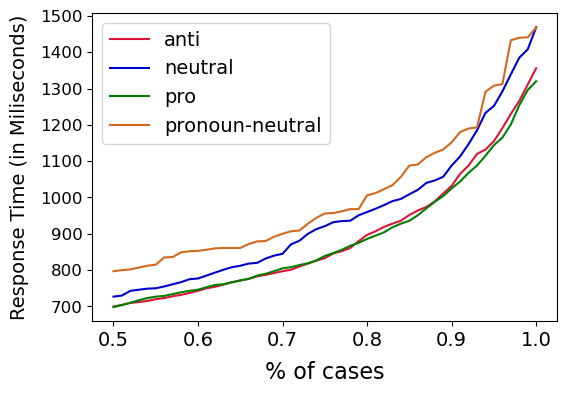}
  \caption{Wino}
  \label{fig:MAZE-thresh-wino}
\end{subfigure}
\begin{subfigure}{0.32\textwidth}
  \includegraphics[width=\textwidth]{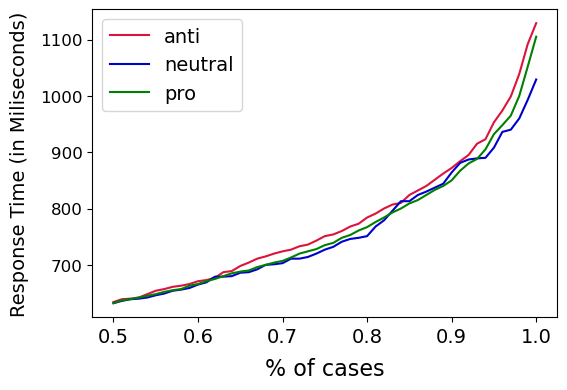}
  \caption{BUG}
  \label{fig:MAZE-thresh-BUG}
\end{subfigure}
\begin{subfigure}{0.32\textwidth}
  \includegraphics[width=\textwidth]{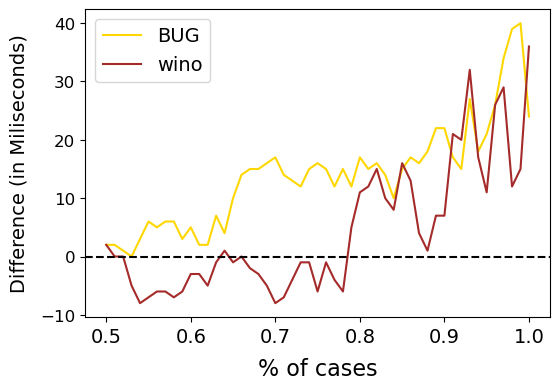}
  \caption{delta-stereotype (anti minus pro)}
  \label{fig:MAZE-delta}
\end{subfigure}
\caption{Figures~\ref{fig:MAZE-thresh-wino} and~\ref{fig:MAZE-thresh-BUG} show the CDF of response times needed to distinguish the pronoun from its distractor in MAZE. I.e., coordinate $(x,y)$ on the graph implies that $x$\% of the annotations required a response time of $y$ ms or less.
Figure~\ref{fig:MAZE-delta} shows $\Delta_{MAZE}$ for humans, i.e., the difference between \as{} response time and \ps{} response time, where values above $y = 0$ indicate gender biased performance.
}
\label{fig:MAZE-thresholds}
\end{figure*}

In this section we summarize the main findings from our two crowdsourcing experiments.
We find that the overall human accuracy for both tasks was good, reaching $94.48\%$ on the gender questions in unrestricted QA, and $98.13\%$ in MAZE, indicating an understandable task and high quality annotations.




\subsection{Experimental Setup}
The two experiments collected annotations from 33 participants on the Amazon Mechanical Turk platform. Our average hourly pay was $8.53$ USD. The overall cost to produce our annotations was $1,030$ USD. To qualify,  workers had to have at least $5,000$ accepted HITs at an acceptance rate of at least $96\%$, and hail from English-speaking countries. In addition, we ran a qualification HIT which required workers to score at least $85\%$ on an unconstrained version of the QA task.  Following \cite{von2020implicit}, we annotated $3K$ instances with gender bias signal for each experiment and each dataset, amounting to  $12K$ annotations. 
We deploy the QA task using Anvil,\footnote{~\url{https://anvil.works/}} and the MAZE task using Ibex.\footnote{~\url{https://github.com/addrummond/ibex}}
Finally, we use the IQR technique to remove outliers in the self-paced reading~\cite{vinutha2018detection}, which may arise due to network connectivity issues. 

\subsection{Evaluation Metrics}

Following previous work, we compute gender bias as the difference in performance between pro-stereotypical and anti-stereotypical instances~\cite{stanovsky-etal-2019-evaluating}. In the QA task we denote as $\Delta_{QA}$  the difference between accuracy on pro-stereotypical versus anti-stereotypical gender questions, which is a proxy for constrained \sys{2} gender bias. In the self-paced reading task we compute the difference in response time to identify the pronoun, marked as $\Delta_{MAZE}$, and is a proxy for \sys{1} biases.
For consistency, both metrics are defined such that larger values indicate more gender biased performance. I.e., for $\Delta_{MAZE}$ we subtract the response time for \ps{} instances from the \as{} instances, as longer response times indicate worse performance.

\subsection{QA results}\label{sec:QA-results}
Several observations can be drawn from the results for the QA task, presented in Table~\ref{tab:QA-results} and visualized in Figure~\ref{fig:QA-results},
showing the biases caused by limiting the resources of \sys{2}.



\textbf{Human subjects show more gender bias as they are given less time to read the sentence.}
For both natural and synthetic sentences, we find that $\Delta_{QA}$ for humans increases between when they are given $0.75$ and $0.5$ of their baseline reading pace, and for natural sentences specifically we see this increase also between $0.5$ and $0.25$. I.e., the difference in performance between \ps{} and \as{} increases the less time participants have. However at some point, participants will not have enough time to process the sentence. This is observed in \winogender{} and \winobias{} when $\alpha = 0.25$, where human performance equally degrades across both  \as{} and \ps{}, in parallel with an increase in non-coreference errors from around $2\%$ when $\alpha\in[0.5,0.75]$ to $5\%$ when $\alpha=0.25$. 


\textbf{Human subjects were found more prone to gender biased answers on naturally-occurring sentences.}
Table~\ref{tab:QA-results} shows larger $\Delta_{QA}$ for natural sentences than for the synthetic ones, and in Figure~\ref{fig:QA-results} the points representing human performance on BUG are farther from the diagonal, indicating more biased performance. This may stem from the templated nature of the synthetic sentences which allows subjects to master them.

\subsection{Self-Paced Reading Results}\label{sec:MAZE-results}

Several conclusions are drawn from this results of this experiment, shown in Figure~\ref{fig:MAZE-thresholds}, approximating \sys{1} biases.

\textbf{Higher human gender bias is observed the more processing time is needed.} Figures~\ref{fig:MAZE-thresh-wino} and~\ref{fig:MAZE-thresh-BUG} show the CDF of response times for distinguishing the pronoun from a distractor over the correct annotations (which consist of 98\% of all annotations). Figure~\ref{fig:MAZE-delta} shows $\Delta_{MAZE}$, which is the difference between \as{} and \ps{} instances in~\ref{fig:MAZE-thresh-wino} and~\ref{fig:MAZE-thresh-BUG}. The longer the response time allowed (and hence more annotations are counted), a more pronounced $\Delta_{MAZE}$ is observed. 


\textbf{Human gender bias was observed only when accounting for at least $80\%$ of the synthetic sentences.}
Positive $\Delta_{MAZE}$ indicates longer response time for \as{} sentences than \ps{}, and so considered gender bias. Figure~\ref{fig:MAZE-delta} shows that $\Delta_{MAZE}$ is positive after accumulating 80\% of the annotations on the CDF curve, while for the natural sentences this effect is found after 50\% of annotations. 



\section{Comparing Model and Human Biases: Discussion and Conclusions}
\label{sec:case-study}

We evaluate  SpanBERT~\cite{joshi-etal-2020-spanbert} and \kirst{}~\cite{kirstain-etal-2021-coreference} on the same sentences annotated by humans in each of the tasks, and compare the bias in results between humans and models.
Below we outline several key findings.

\begin{figure*}
\centering
\begin{subfigure}{0.85\columnwidth}
  \includegraphics[width=0.95\columnwidth]{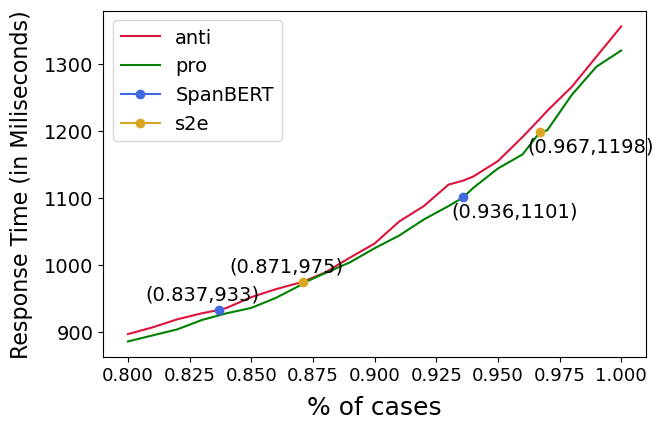}
  \caption{Wino}
  \label{fig:MAZE-models-wino}
\end{subfigure}\hspace{12mm}
\begin{subfigure}{0.85\columnwidth}
  \includegraphics[width=0.95\columnwidth]{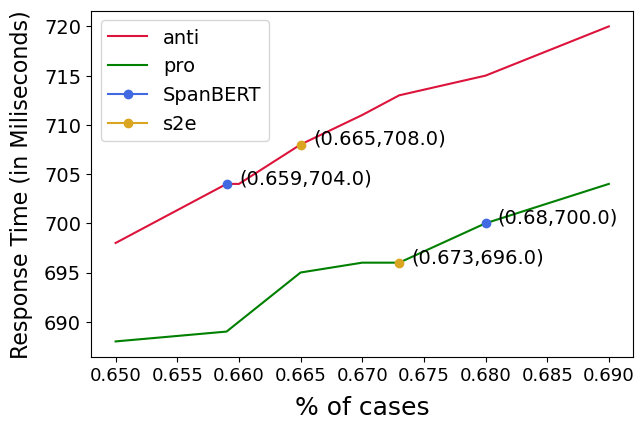}
  \caption{BUG}
  \label{fig:MAZE-models-BUG}
\end{subfigure}
\caption{Model performance versus human annotations. The blue and yellow points are the intersection points with the different models' accuracy and their matching category threshold. For example, the blue point intersecting the red line, is the human threshold that matches SpanBERT accuracy on \as{} sentences.}
\label{fig:MAZE-models}
\end{figure*}

\begin{figure}[tb!]
\centering
\includegraphics[width=0.92\columnwidth]{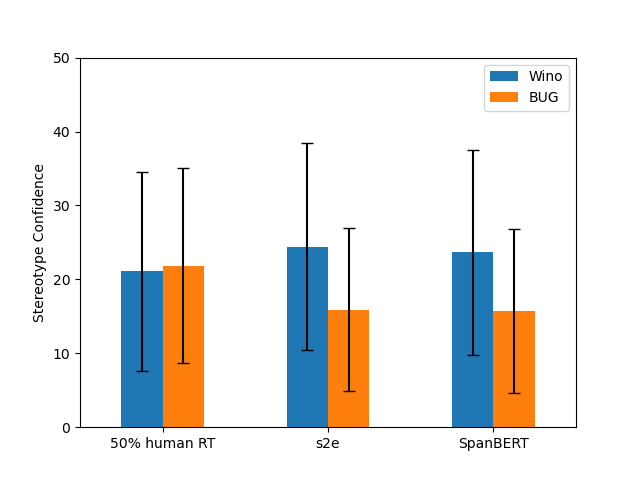}
\caption{The mean and std of ``stereotype confidence" for humans versus models errors, defined as the average distance from 50\% gender distribution of profession, according to the U.S. Bureau of Labor.} 
\label{fig:QA-qualitative}
\end{figure}

\textbf{Qualitative error analysis.}
In Figure~\ref{fig:QA-qualitative} we compare errors made by human and models, and find that models tend to err on professions which are strongly associated with a specific gender according the U.S. Bureau of Labor Statistics, while humans err more broadly, on less stereotypical assignments.

\textbf{Models exhibit gender bias more than humans  on synthetic sentences in the QA experiment.}
Table~\ref{tab:QA-results} shows that $\Delta_{QA}$ on \winogender{} and \winobias{} is larger for models when compared to humans on any fraction of the reading pace. Additionally, Figure~\ref{fig:QA-results} shows that over \winogender{} and \winobias{}, models are farther from equilibrium line than humans, and the human performance on \as{} instances is superior to models. 
This may indicate that to achieve good performance, models rely on gender bias more than humans.

\textbf{Models show more gender bias on synthetic sentences than in real-world sentences, as opposed to humans where gender bias is more pronounced over natural sentences.} 
For the QA experiment this trend is seen in $\Delta_{QA}$ columns in Table~\ref{tab:QA-results}. As for the self-paced reading task, Figure~\ref{fig:MAZE-delta} shows that human's $\Delta_{MAZE}$ on BUG is above $\Delta_{MAZE}$ on \winogender{} and \winobias{}, while for models $\Delta_{MAZE}$ is the distance on the x-axis between the points in Figures~\ref{fig:MAZE-models-wino} and \ref{fig:MAZE-models-BUG}, which is smaller on BUG for both models.  
In humans, this may arise due to mastering synthetic sentences to the point they do not rely on gender stereotypes to excel in it. 
In contrast, the degraded performance of models on real-world sentences diminishes the gains from biased predictions.

\textbf{Models present higher accuracy on the subset of sentences used for the self-paced reading experiment than on the subset used for the QA experiment for both datasets.}
We found that model accuracy is higher on the subset of sentences used in the self-paced reading experiment. In average, between \winobias{} and \winogender{} subsets we see $8.05\%$ better performance on the self-paced reading subset, and between BUG subsets we see $5.4\%$ better performance.
This may indicate that models do better on sentences where the pronoun appears after verb, as is the case in our self-paced reading experiment, detailed in the Experiments section.

\textbf{Conclusion: Model biases reflect human decision-making under constrained settings.}
Revisiting our research question, our findings suggest that gender bias in coreference resolution is comparable to human biases rather than an annotation artifact, indicating it will likely creep up in real-world datasets along with other, more desired human behavior, like common sense reasoning. 

\textbf{Future Work.}
Follow-up work may compare our results with competing  cognitive theories, e.g., ~\cite{bursell2021we}, as well as developing some kind of ``slow reasoning'' models, e.g., via early exiting~\cite{Schwartz2020TheRT,Laskaridis2021AdaptiveIT}.

\section{Limitations}
As with any work involving humans subjects in general, and crowdsourcing in particular, several limitations might arise. 
First, crowdsourcing results are less reliable than a controlled in-house experiment. Second, the conclusions we derived may apply only for our group of annotators. In addition, we did not use any demographic details regarding our participants so our results may be prone to societal biases and not represent the phenomenon for other real-world distributions. To address this, future work can validate our results over larger and more diverse annotator cohorts.

A linguistic limitation of our work is that we only refer to the observed phenomena in English. Our work can only be generalized to languages without gender inflection nouns, as our proposed methodology assumes that the gender of the profession is obtained through the pronoun referring to it. 

Another limitation of our work is that we only address pronouns that are grammatically feminine, masculine or neutral. We did not address pronouns that can match to more than one of them (``his/her''), or pronouns that match other gender identities.

Finally, a societal limitation of our work is that our definitions for \as{} and \ps{} are done according to U.S.-centric societal norms which may diverge between cultures.

\newpage
\section{Acknowledgments}
We thank Yevgeni Berzak and Roger Levy for many fruitful discussions, and the anonymous reviewers for their valuable feedback. This work
was supported in part by a research gift from the Allen Institute for AI, and a research grant 2336 from the Israeli Ministry of Science and Technology.




\newpage
\section{Appendix}
\subsection{Datasets Analysis}

In this work we thoroughly investigate relations between pronouns and entities that represented by professions. In the following table we present all pronouns that appear in our corpora, along with their distribution, to better understand what affects the examined trends.
\begin{table}[ht]
\centering
\resizebox{\columnwidth}{!}{%
\begin{tabular}{@{}llllclll@{}}
\toprule
                           &                              & \multicolumn{2}{c}{\winogender} & \multicolumn{2}{c}{\winobias}     & \multicolumn{2}{c}{BUG} \\ \midrule
\multirow{4}{*}{masculine} & \multicolumn{1}{l|}{he}      & 178          & (74\%)          & \multicolumn{1}{l}{731} & (46\%) & 423       & (32\%)      \\
                           & \multicolumn{1}{l|}{his}     & 54           & (23\%)          & \multicolumn{1}{l}{149} & (9\%)  & 742       & (55\%)      \\
                           & \multicolumn{1}{l|}{him}     & 8            & (3\%)           & \multicolumn{1}{l}{706} & (45\%) & 23        & (2\%)       \\
                           & \multicolumn{1}{l|}{himself} & \multicolumn{2}{c}{-}          & \multicolumn{2}{c}{-}            & 149       & (11\%)      \\ \midrule
\multirow{3}{*}{feminine}  & \multicolumn{1}{l|}{she}     & 178          & (74\%)          & \multicolumn{1}{l}{700} & (44\%) & 132       & (34\%)      \\
                           & \multicolumn{1}{l|}{her}     & 62           & (26\%)          & \multicolumn{1}{l}{882} & (56\%) & 221       & (58\%)      \\
                           & \multicolumn{1}{l|}{herself} & \multicolumn{2}{c}{-}          & \multicolumn{2}{c}{-}            & 30        & (8\%)       \\ \midrule
\multirow{3}{*}{neutral}   & \multicolumn{1}{l|}{they}    & 178          & (74\%)          & \multicolumn{2}{c}{-}            & \multicolumn{2}{c}{-}   \\
                           & \multicolumn{1}{l|}{their}   & 54           & (23\%)          & \multicolumn{2}{c}{-}            & \multicolumn{2}{c}{-}   \\
                           & \multicolumn{1}{l|}{them}    & 8            & (3\%)           & \multicolumn{2}{c}{-}            & \multicolumn{2}{c}{-}   \\ \bottomrule
\end{tabular}%
}
\caption{Distribution of grammatical gender pronouns that appear in sentences in original corpora.}
\label{tab:grammatical-gender-pronouns}
\end{table}

\subsection{Fine Grained Categorization}
Table~\ref{tab:wino-types} presents our suggested fine grained categorization, that takes in consideration also the societal norms of the other entity profession, in addition to the original datasets labeling, that only looks at the societal norms regarding the main entity.

\begin{table*}
\centering
\resizebox{0.97\textwidth}{!}{%
\begin{tabular}{@{}lllll@{}}
\toprule
\textbf{Type} &
  \textbf{Explanation} &
  \textbf{Example} &
  \textbf{\#sentences} &
  \textbf{\% in data} \\ \midrule
\cellcolor[HTML]{F6B0B0} 
anti-pro &
  \begin{tabular}[c]{@{}l@{}}The main entity's occupation does not match the \\ pronoun gender, but other entity's occupation matches.\end{tabular} &
  \begin{tabular}[c]{@{}l@{}}The sheriff questioned the housekeeper as she \\ needed to find out the thief.\end{tabular} &
  1586 &
  40.7\% \vspace{0.5mm}\\\hline
\cellcolor[HTML]{C0F6B7} 
pro-anti &
  \begin{tabular}[c]{@{}l@{}}The main entity occupation matches the pronoun\\ gender, but other entity does not match.\end{tabular} &
  \begin{tabular}[c]{@{}l@{}}The supervisor helped the writer and then\\ asked her to return the favor.\end{tabular} &
  1582 &
  40.7\% \vspace{0.5mm}\\\hline
\cellcolor[HTML]{D3D3D3} 
pronoun-neutral &
  The pronoun in the sentence is neutral. &
  \begin{tabular}[c]{@{}l@{}}The appraiser told someone that they had paid \\ too much for the painting.\end{tabular} &
  240 &
  6.2\% \vspace{0.5mm}\\\hline
\cellcolor[HTML]{F6B0B0} 
anti-neutral &
  \begin{tabular}[c]{@{}l@{}}The main entity occupation does not match the pronoun \\ gender, other entity's occupation is considered neutral.\end{tabular} &
  \begin{tabular}[c]{@{}l@{}}The homeowner asked the inspector if she \\ had discovered any building code violations.\end{tabular} &
  108 &
  2.8\% \vspace{0.5mm}\\\hline
\cellcolor[HTML]{C0F6B7} 
pro-neutral &
  \begin{tabular}[c]{@{}l@{}}The main entity occupation matches the pronoun\\ gender, other entity's occupation is considered neutral.\end{tabular} &
  \begin{tabular}[c]{@{}l@{}}The physician warned someone that he could\\ not safely prescribe a higher dose.\end{tabular} &
  108 &
  2.8\% \vspace{0.5mm}\\\hline
\cellcolor[HTML]{C0F6B7} 
neutral-anti &
  \begin{tabular}[c]{@{}l@{}}The main entity's occupation is considered neutral, other\\ entity's occupation does not match the pronoun gender.\end{tabular} &
  \begin{tabular}[c]{@{}l@{}}The instructor encouraged someone to pursue \\ his dreams.\end{tabular} &
  108 &
  2.8\% \vspace{0.5mm}\\\hline
\cellcolor[HTML]{F6B0B0} 
neutral-pro &
  \begin{tabular}[c]{@{}l@{}}The main entity's occupation is considered neutral, and\\ other entity's occupation matches the pronoun gender.\end{tabular} &
  \begin{tabular}[c]{@{}l@{}}The teenager confided in the therapist because \\ she was seeking emotional support.\end{tabular} &
  108 &
  2.8\% \vspace{0.5mm}\\\hline
\cellcolor[HTML]{D3D3D3} 
neutral-neutral &
  \begin{tabular}[c]{@{}l@{}}Both main and other entity's occupations are \\ considered neutral.\end{tabular} &
  \begin{tabular}[c]{@{}l@{}}The worker told the pedestrian that he was\\ repairing the sidewalk as quickly as possible.\end{tabular} &
  48 &
  1.2\% \\ \bottomrule
\end{tabular}%
}
\caption{Our suggested fine grained categorization of \winogender{} and \winobias{} sentences. The first column identifies the stereotypical assignment for each of the entities. Background colors stand for our fine-grained categorization,  green stands for \ps{}, red for \as{} and gray for neutral. Additionally to the original labeling schema, our categorization is determined also by the gender stereotype of the \textbf{other} entity, as exemplified in lines labeled \emph{neutral-anti} and \emph{neutral-pro}.
}
\label{tab:wino-types}
\end{table*}




\subsection{Human Annotators}
\label{sec:human-annotators}
\begin{figure*}[htbp!]
    \centering
    \includegraphics[width=\textwidth]{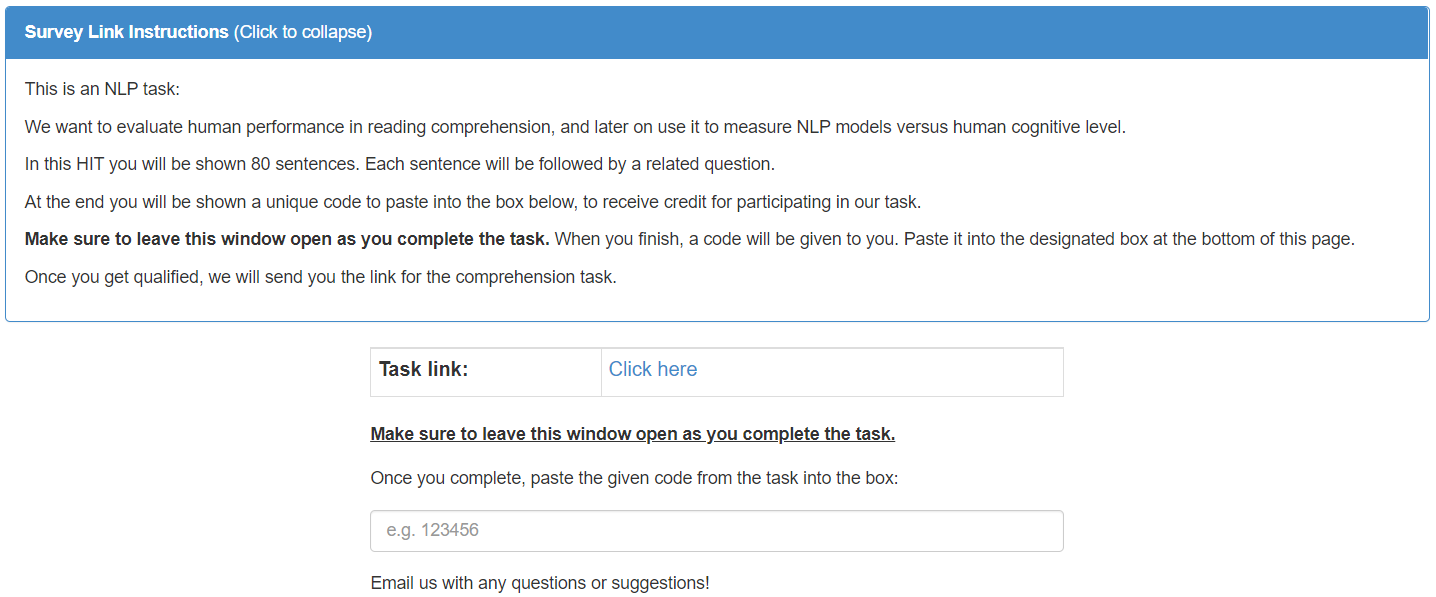}
    \caption{Instructions for our QA qualification task. 
    }
    \label{fig:instructions}
\end{figure*}

Figure~\ref{fig:instructions} is an example for the first window in each of our experiments. We show instructions for the coming task, and users actively press the link as a consent for participation. This template of instructions was shown in each of our experiments.

\subsection{Changelog: CogSci2023 to Current Version}
This paper is an extended version of a paper appearing in CogSci 2023. Here is the list of the sections we added:
\begin{itemize}
    \item Figure 1
    \item Sample Instances for Human Annotation (section)
    \item Qualitative error analysis (paragraph+Figure 7)
    \item Models present higher accuracy on the subset of sentences used for the self-paced reading experiment than on the subset used for the QA experiment for both datasets (paragraph)
    \item Limitations (section)
    \item Appendix
\end{itemize}

\end{document}